# A novel multiclassSVM based framework to classify lithology from well logs: a real-world application

Soumi Chaki[1], Aurobinda Routray[1], William K. Mohanty[2], Mamata Jenamani[3]

[1]Department of Electrical Engineering,
IIT Kharagpur, India
soumibesu2008@gmail.com
aroutray@ee.iitkgp.ernet.in

[2] Department of Geology and Geophysics,
IIT Kharagpur, India
wkmohanty@gg.iitkgp.ernet.in

[3]Department of Industrial and Systems Engineering,
IIT Kharagpur, India
mj@iem.iitkgp.ernet.in

*Abstract*— Support vector machines (SVMs) have been recognized as a potential tool for supervised classification analyses in different domains of research. In essence, SVM is a binary classifier. Therefore, in case of a multiclass problem, the problem is divided into a series of binary problems which are solved by binary classifiers, and finally the classification results are combined following either the one-against-one or one-against-all strategies. In this paper, an attempt has been made to classify lithology using a multiclass SVM based framework using well logs as predictor variables. Here, the lithology is classified into four classes such as sand, shaly sand, sandy shale and shale based on the relative values of sand and shale fractions as suggested by an expert geologist. The available dataset consisting well logs (gamma ray, neutron porosity, density, and P-sonic) and class information from four closely spaced wells from an onshore hydrocarbon field is divided into training and testing sets. We have used one-against-all strategy to combine the results of multiple binary classifiers. The reported results established the superiority of multiclass SVM compared to other classifiers in terms of classification accuracy. The selection of kernel function and associated parameters has also been investigated here. It can be envisaged from the results achieved in this study that the proposed framework based on multiclass SVM can further be used to solve classification problems. In future research endeavor, seismic attributes can be introduced in the framework to classify the lithology throughout a study area from seismic inputs.

## I. Introduction

Reservoir characterization includes prediction [1], [2] or classification [3], [4] of reservoir characteristics, mainly porosity, permeability, saturation, lithology, pay zone thickness, reservoir pressure and temperature etc. from available data. The existing software supports available to handle uncertainties and inexactness of the measurements may not be adequate in all cases. Therefore, more sophisticated soft computing tools are needed to be applied for increasing the accuracy of reservoir characterization. Due to nonlinear and heterogeneous nature of reservoir variables, it is very difficult to model such characteristics using simple regression and also from input/output neural network structure. Therefore, it is argued that classification based modeling of reservoir properties can achieve the modeling objectives accurately. Different classifiers such as naïve Bayes, discriminant, and neural network based classifiers are used in different classification problems. In case of imbalanced dataset, support vector data description has also been used to identify minority class patterns in the testing set [5]. Literature study reveals that SVM has been used to solve classification problems in different domains of research such as hyperspectral imaging [6], reservoir characterization [7], imbalanced dataset [8], and machines [9]. The theory and approaches of SVM (binary/multiclass) has been discussed in length [10]–[13].

SVM is fundamentally a binary classifier. However, in case of multiclass problems, the problem can be decomposed into several binary class problems solvable by a binary classifier; then, either the one-against-one or one-against-all strategy is followed to combine to obtain the solution. In this paper, we have used a multiclass SVM based framework to classify lithology from well logs. For this work, the identification of the lithology classes has been carried out from values of sand and shale fractions as suggested by expert geologists. In this study, the number of patterns has been much more compared to the number of predictor attributes and the number of patterns belonging to the different classes are comparable; therefore we have envisaged that multiclass SVM would work satisfactorily.

The contributions of the study can be listed as:

- A complete classification framework
- Comparison with other classifiers
- Investigation on the available predictors
- The effect of different kernel and parameter selection on the performance of multiclass SVM

The rest of the paper is structured as follows: first, the fundamentals of SVM classification are discussed before presentations of the dataset used in this study. Then, the proposed classification framework is presented. The next section provided the results achieved in this study and a discussion before concluding the paper along with future scope in the following section.

## II. SUPPORT VECTOR MACHINE (SVM)

Mostly the Support Vector Machines have been widely considered as a two-class classifier. To understand the basic concept of SVM classifier, first, let us consider, a training sample $\{(x_i, d_i)\}_{i=1}^{N}$, where $x_i$ and $d_i$ represent the input and corresponding desired response of the $i^{th}$ example pattern respectively. The class represented by the subset $d_i = +1$ and $d_i = -1$ are "linearly separable" [14]. The hyperplane equation that satisfies the separation is

$$w^T x + b = 0 \qquad (1)$$

where $x$ is an input vector, $w$ is an adjustable weight vector, and $b$ is a bias. Therefore, this equation can be re-written as,

$$w^T x_i + b \geq 0 \text{ for } d_i = +1$$
$$w^T x_i + b < 0 \text{ for } d_i = -1 \qquad (2)$$

The separation between the hyperplane defined in Eq. (7) and the closest data point is called *margin of separation* for a given weight vector and bias. The goal of a support vector machine is to find the particular hyperplane to maximize the margin of separation, $\rho$. Under this condition, the decision surface is referred to as the *optimal hyperplane*. If $w_0$ and $b_0$ are the optimum values of the weight vector and bias respectively, then the corresponding *optimal hyperplane* is defined by

$$w_0^T x + b_0 = 0 \qquad (3)$$

and the *discriminant function* given as:

$$g(x) = w_0^T x + b_0 \qquad (4)$$

It is an algebraic measure of the distance from $x$ to the hyperplane. The expression can be re-written as

$$x = x_p + r \frac{w_o}{\|w_o\|} \qquad (5)$$

where $x_p$ is the normal projection of $x$ onto the hyperplane and $r$ is the desired algebraic distance. The distance $r$ is positive or negative depending on the placement of $x$ on the positive or negative side of the hyperplane. The equation of *discriminant function* can be represented as

$$g(x) = w_0^T x + b_0 = r \|w_0\| \qquad (6)$$

Let us consider a support vector $x^{(s)}$ for which $d^{(s)} = 1$. Then it can be written as,

$$g(x^{(s)}) = w_0^T x^{(s)} + b_0 = \mp 1 \text{ for } d^{(s)} = \mp 1 \qquad (7)$$

The algebraic distance from the support vector $x^{(s)}$ to the *optimal hyperplane* can be written as,

$$r = \frac{g(x^{(s)})}{\|w_0\|}$$
$$= \frac{1}{\|w_0\|} \text{ if } d^{(s)} = +1 \qquad (8)$$
$$= -\frac{1}{\|w_0\|} \text{ if } d^{(s)} = -1$$

where the plus and minus sign indicate that $x^{(s)}$ lie on the positive and negative side of the hyperplane respectively. If $\rho$ denotes the optimum value of the *margin of separation* between the two classes from the training sample then, it can be represented as

$$\rho = 2r = \frac{2}{\|w_0\|} \qquad (9)$$

Therefore, the optimum condition is achieved by minimizing the Euclidian norm of the weight vector $w$.

The weight vector $w$ minimizes the cost function:

$$\phi(w) = \frac{1}{2} w^T w \qquad (10)$$

Subject to the constraint

$$d_i(w^T x_i + b) \geq 1 \text{ for } i = 1, 2, ..., N \qquad (11)$$

We construct the Lagrangian function to solve the constrained optimization problem.

$$J(w, b, \alpha) = \frac{1}{2} w^T w - \sum_{i=1}^{N} \alpha_i [d_i(w^T x_i + b) - 1] \qquad (12)$$

Where $\alpha_i$ are called Lagrangian multipliers. The solution can be represented by the saddle point of the Lagrangian function which has to be minimized with respect to $w, b$ and maximized with respect to $\alpha$.

Differentiating the Lagrangian function with respect to $w, b$ and making it to zero yields the conditions

$$w = \sum_{i=1}^{N} \alpha_i d_i x_i \qquad (13)$$

$$\sum_{i=1}^{N} \alpha_i d_i = 0 \qquad (14)$$

Further utilizing the Kuhn–Tucker conditions,

$$\sum_{i=1}^{N} \alpha_i [d_i(w^T x_i + b) - 1] = 0 \text{ for } i = 1, 2, ..., N \qquad (15)$$

The optimization in the dual domain yields the convex solution of the weights and bias for the hyperplane.

## III. DESCRIPTION OF THE DATASETS

The study area is located at an onshore hydrocarbon field ate western part of India. The logs are acquired from four closely spaced wells to be referred as A, B, C, and D, respectively.

The dataset contains several logs such as gamma ray content (GR), bulk density (RHOB), P-sonic (DT), neutron porosity (NPHI) and reservoir characteristics, e.g., sand fraction and shale fraction. In recent literature, gamma ray content (GR), bulk density (RHOB), P-sonic (DT), neutron porosity (NPHI), spontaneous potential (SP) are used to model or classify lithological properties [4], [15]. The detailed description of the logs used as predictors in this study can be found in [4].

The classification of lithology based on the values of sand ($v_{sand}$) and shale ($v_{shale}$) fractions has been carried out by experts' opinion. The range of the sand and shale fractions vary between zeros to one.

TABLE I. CLASSIFICATION OF LITHOLOGY

| Criteria | Lithology |
|---|---|
| Sand | $v_{shale} \leq 0.15$ |
| Shaly sand | $0.15 < v_{shale} \leq 0.5$ and $v_{sand} > v_{shale}$ |
| Sandy shale | $0.5 < v_{shale} \leq 0.65$ and $v_{sand} < v_{shale}$ |
| Shale | $v_{shale} > 0.65$ and $v_{sand} < v_{shale}$ |

Table I represents the criteria of classification of lithology given the sand and shale fractions. It can be seen from Table I that the lithology of the study area can be classified into four classes such as sand, shaly sand, sandy shale, and shale depending on the sand and shale fractions.

## IV. PROPOSED FRAMEWORK

The complete framework used in this study can be divided into three stages: data collection, pre-processing, and finally training and testing. The workflow is implemented on a 64-bit MATLAB platform installed in on a Intel(R) Core(TM) i5 CPU @3.20 GHz workstation having 16 GB RAM.

### A. Data Collection

In the first stage as shown in Fig. 1, the data acquisition is carried out. The collected logs reveal the presence of a number of missing values. The samples consisting missing values are removed from the data file to produce a processed file ready to be used for analysis.

### B. Pre-processing

This is the second stage in the proposed framework. Pre-processing plays a crucial role to obtain satisfactory performance of the designed framework. Literature shown application of elegant pre-processing steps to improve the performance of prediction and classification algorithms in reservoir characterization domain such as [16], [17].

The pre-processing stage involves re-sampling, normalization, and exploring relevance of the predictor attributes used in this study. The dataset containing predictor variables GR, RHOB, DT, NPHI, and class labels decided based on the sand and shale fractions following Table I are used for pre-processing. First, the samples are uniformly re-sampled; the resulting sampling interval of the dataset is now 0.15 milliseconds. In the next step, the predictors are normalized using data of all wells taken together by z-score normalization as in [15]. In this stage, the relevance of the predictors has been investigated using Relieff algorithm as in [4].

### C. Training and testing

The task of training and testing is implemented to evaluate the performances of the multiclass SVM based classifier. Here, the normalized dataset along with the class information is used. First, the training dataset is created by combining 70% sample patterns from each of the wells and the patterns are scrambled to remove any trend along the depths. The remaining 30% samples from each of the four wells are combined and scrambled to make the testing set to verify the performance of the trained classifier.

The performance of multiclass SVM is improved by adjusting few parameters: type of the kernel function and parameters associated with selected kernel. After training, the trained classifier is validated using the testing patterns and the performance is evaluated in terms of classification accuracy from the resulted confusion matrix using (16). Here, $TP, TN$ represent true positive and true negatives which are the diagonal elements of the confusion matrix and $FP, FN$ represent the false positives and false negatives represented by the off-diagonal elements of the resulted confusion matrix.

$$\text{accuracy} = \frac{TP + TN}{FP + FN} \quad (16)$$

After completion of the training and testing, the classification performance achieved using this proposed framework has been compared to other supervised classifiers namely naive Bayes, and artificial neural network (ANN). Moreover, the performance of kernels and associated parameters along with effect of different combinations of predictor variables have been investigated in this study.

## V. EXPERIMENTAL RESULTS

The experimental results and analysis carried out in this study are reported in this section. Table II represents the statistics of the predictors in terms of maximum, minimum, mean and standard deviation combining dataset of the four wells taken together. The difference among the range of the predictors has necessitated the normalization step before the classifier training.

TABLE II. STATISTICS OF THE PREDICTOR VARIABLES

| Predictors | Maximum | Minimum | Mean | Standard Deviation |
|---|---|---|---|---|
| GR | 104.82 | 13.45 | 57.93 | 18.07 |
| NPHI | 0.61 | 0.09 | 0.35 | 0.09 |
| RHOB | 2.86 | 1.69 | 2.45 | 0.11 |
| DT | 138.24 | 60.68 | 96.73 | 11.76 |

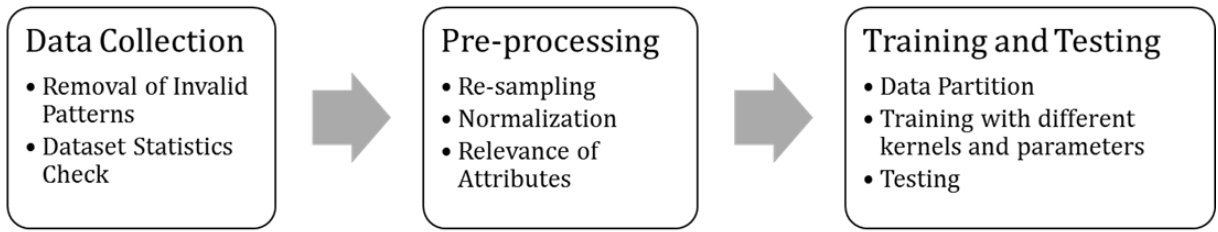

Fig. 1. Complete proposed framework including data collection, pre-processing, and training and testing.

Fig. 2 represents the distribution of the predictor variables along depth for well A. Fig. 3 represents the relevance of predictor attributes using Relief algorithm in terms of predictor importance weight. It demonstrates that the importance of P-sonic log is less compared to its three other counterparts for lithology classification. The dependency of this revelation can be verified later.

The calibration of the classifier has been carried out after normalization of the attributes with different combinations of kernel function and associated parameters. Finally, the radial basis kernel function (rbf kernel) with sigma value of 0.5 has yielded best results during testing. Then, the performance of the SVM classifier is compared to other supervised classifiers namely naive Bayes, and artificial neural network (ANN) based classifiers which are optimized after initializing with appropriate parameter values using the same set of predictors. Table II represents the result of the comparison in terms of classification accuracy. It is evident from Table II that the multiclass SVM based framework outperformed the other classifiers. The classification accuracy of the multiclass SVM with rbf kernel having sigma 0.5 has been highlighted in Table III.

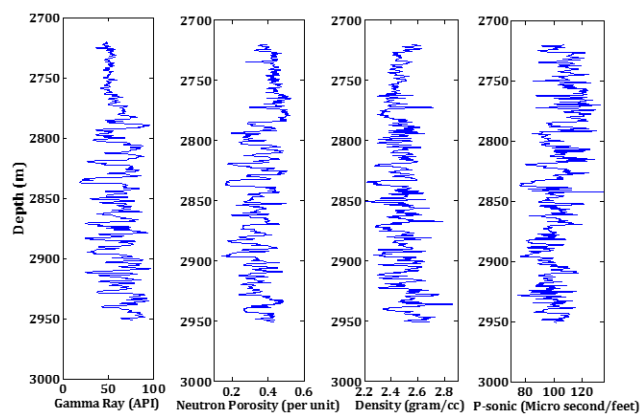

Fig. 2. Predictor variables along depth for well A

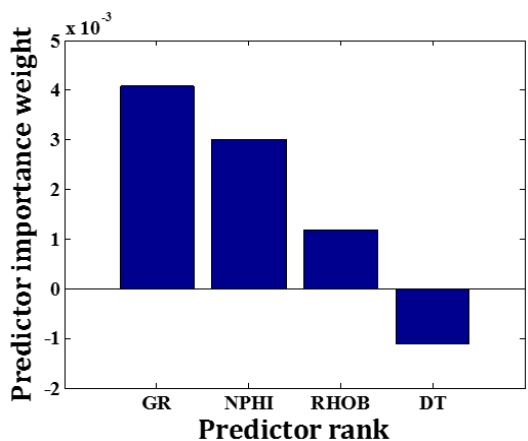

Fig. 3. Relevance of predictor attributes using Relief algorithm in terms of predictor importance weight

TABLE III. PERFORMANCE COMPARISON OF CLASSIFIERS

| Classifier | | Classification accuracy (%) |
|---|---|---|
| *Multiclass SVM* | Linear kernel | 66.82 |
| | rbf kernel | **82.43** |
| *Naïve Bayes* | | 71.25 |
| *ANN* | | 75.1 |

Fig. 4 represents confusion matrix plot for testing patterns after calibrating a multiclass SVM classifier with rbf kernel, sigma=0.5 using the training parameters. Here, the diagonal elements of the matrix represent the true positive values corresponding to the respective class. For example, 0.98 in the first row represents that the 98% of the patterns belong to sand class is correctly classified and only 2% pattern are misclassified as shaly sand.

|  | Sand | Shaly sand | Sandy shale | Shale |
|---|---|---|---|---|
| Sand | 0.98 | 0.02 | 0.00 | 0.00 |
| Shaly sand | 0.22 | 0.75 | 0.02 | 0.01 |
| Sandy shale | 0.00 | 0.08 | 0.87 | 0.05 |
| Shale | 0.00 | 0.00 | 0.34 | 0.65 |

Fig. 4. Confusion matrix achieved by trained multiclass SVM with rbf kernel, sigma=0.5

It can be observed from Fig. 4 that among the four classes, the amount of misclassification is more in case of shale class. The predictor variables are unable to firmly distinguish between sandy shale and shale classes. Similarly, there are overlaps while distinguishing between sand and shaly sand classes. However, the classification result is acceptable as the overlaps are taking place only in case of adjacent classes. There is no overlap between sand and sandy shale classes as well as between sand and shale classes.

As stated above the relevance of the predictor has been verified by training the classifier with different sets of inputs. According to Fig. 3, the relative importance of GR, NPHI are more compared to RHOB and DT. Table IV represents performance of the classifier trained by the proposed multiclass SVM with different combinations of inputs. In all cases, the rbf kernel has been used with sigma value 0.5. It is evident from Table IV that though RHOB, DT have less importance compared to GR and NPHI; still these two predictors has improved the classification accuracy. Hence, all the four predictors have been used to build and test the classification model for lithology.

TABLE IV. PERFORMANCE COMPARISON OF CLASSIFIERS

| Classifier | Classification accuracy (%) |
|---|---|
| GR+NPHI | 65.07 |
| GR+NPHI+RHOB | 73.72 |
| GR+NPHI+RHOB+DT | 82.43 |

The analysis has revealed that the performance of the multiclass SVM also dependent on the selection of appropriate parameters for a specific kernel function. Fig. 5 represents the variation in classification accuracy with respect to variation in the sigma values (0 to 2.0) while training with rbf kernel. It can be found from Fig. 5 that rbf kernel with 0.5 as sigma value has performed well while using multiclass SVM based classifier.

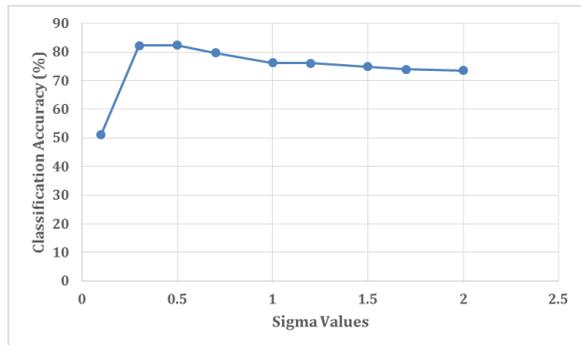

Fig. 5. Variation in classification accuracy while training the classifier using multiclass SVM with rbf kernel with respect to different sigma values

Therefore, it can be inferred from the results that the proposed workflow based on multiclass SVM can be used as a powerful tool to classify lithology of a particular area from several well logs in reservoir characterization domain.

## VI. Conclusions

In this work, a complete framework based on multiclass SVM is proposed to classify lithology from well logs into four classes as suggested by an expert geologists. Comparative analysis reported in this study has established the superiority of the proposed methodology compared to other supervised classification algorithms in terms of accuracy. The performance of the multiclass SVM with different kernel functions and associated parameters has also been investigated. Moreover, each step in the proposed framework is important to achieve overall performance of the framework. The propose work signifies the applicability of the concepts of signal processing and machine learning to solve a real-world problem; thus expanding the scope of interdisciplinary researches.

This work can be extended with inclusion of seismic attributes as inputs in the framework. Integration of seismic and limited number of available borehole data will be helpful to produce area map of the lithology of the study area.